\pdfoutput=1
\documentclass[11pt]{article}
\usepackage{EMNLP2022}
\usepackage{times}
\usepackage{latexsym}
\usepackage{amsmath}
\usepackage[T1]{fontenc}
\usepackage[utf8]{inputenc}
\usepackage{microtype}
\usepackage{inconsolata}
\usepackage{graphicx}
\usepackage{float}
\graphicspath{ {./image/} }
\usepackage{multirow}
\usepackage{booktabs}
\usepackage{dsfont}
\usepackage{arydshln}
\usepackage{enumitem}
\usepackage{CJKutf8}
\usepackage{bm}
\usepackage{amssymb}
\usepackage{hyperref}
\usepackage{url}
\usepackage{color}
\definecolor{darkblue}{rgb}{0, 0, 0.5}
\hypersetup{colorlinks=true, citecolor=darkblue, linkcolor=darkblue, urlcolor=darkblue}

\title{Breaking the Representation Bottleneck of Chinese Characters: \\ Neural Machine Translation with Stroke Sequence Modeling}

\author{Zhijun Wang~~~
        Xuebo Liu\thanks{~~Co-first and Corresponding Author}~~~
        Min Zhang \\
  Institute of Computing and Intelligence, Harbin Institute of Technology, Shenzhen, China\\
  \texttt{1190303311@stu.hit.edu.cn, \{liuxuebo,zhangmin2021\}@hit.edu.cn} \\
}

\begin{document}
\maketitle
\begin{CJK*}{UTF8}{gbsn}
\begin{abstract}
Existing research generally treats Chinese character as a minimum unit for representation.
However, such Chinese character representation will suffer two bottlenecks: 1) Learning bottleneck, the learning cannot benefit from its rich internal features (e.g., radicals and strokes); and 2) Parameter bottleneck, each individual character has to be represented by a unique vector.
In this paper, we introduce a novel representation method for Chinese characters to break the bottlenecks, namely StrokeNet, which represents a Chinese character by a Latinized stroke sequence (e.g., ``凹 (concave)'' to ``ajaie'' and ``凸 (convex)'' to ``aeaqe'').
Specifically, StrokeNet maps each stroke to a specific Latin character, thus allowing similar Chinese characters to have similar Latin representations.
With the introduction of StrokeNet to neural machine translation (NMT), many powerful but not applicable techniques to non-Latin languages (e.g., shared subword vocabulary learning and ciphertext-based data augmentation) can now be perfectly implemented.
Experiments on the widely-used NIST Chinese-English, WMT17 Chinese-English and IWSLT17 Japanese-English NMT tasks show that StrokeNet can provide a significant performance boost over the strong baselines with fewer model parameters, achieving 26.5 BLEU on the WMT17 Chinese-English task which is better than any previously reported results without using monolingual data.
Code and scripts are freely available at \url{https://github.com/zjwang21/StrokeNet}.
\end{abstract}

\section{Introduction} \label{introduction}
Neural machine translation (NMT) has become the mainstream paradigm in machine translation recently \citep{sutskever2014sequence, cho2014learning, bahdanau2014neural}. With rich bilingual parallel corpora, NMT achieves state-of-the-art performance on multiple translation benchmarks. In Chinese NMT tasks, the Chinese character has been the minimum representation unit for a long time. However, such representation perhaps might not be the best choice for Chinese NMT due to the two following representation bottlenecks.

\begin{table}[t]
\centering
\scalebox{0.93}{
\begin{tabular}{ll}
\toprule
Zh &\textcolor{red}{布} 什 和 沙 龙 举 行 \textcolor{blue}{了} 会 谈 \\
En &Bush held a talk with Sharon \\
\hdashline
\multirow{2}{*}{Zh (Stroke)} &\textcolor{red}{etasa} taea teatoaie oodatot etcto   \\ &ootetneea ttaeer \textcolor{blue}{hr} tneelo oyottoottn \\
\bottomrule
\end{tabular}}
\caption{StrokeNet represents a Chinese character by a Latinized stroke sequence. For example, ``布'' to ``etasa'' and ``了'' to ``hr''.}
\label{tab:intro}
\end{table}

The first is the learning bottleneck. The representation learning of Chinese does not fully utilize its rich internal features. Latin languages have rich information in words like affixes. Actually, Chinese also has this kind of internal information. A Chinese character usually contains one radical (rarely has two) and several other radical-like components~\citep{li2015component}. Characters with the radical ``扌'' commonly are verbs. The characters ``扎(tie)'', ``拉(pull)'', ``打(hit)'', ``扔(throw)'' and ``提(carry)'' all have the meaning of acting with hands because they have the same radical ``扌''. Latin languages can easily learn this internal information by subword modeling while Chinese cannot if just taking character as the minimum unit into consideration, which limits the representation capability of NMT models.

The second is the parameter bottleneck.
In Chinese NLP models, the parameters used for Chinese word representation can be a huge number.
In large-scale cross-lingual pre-trained language models like XLM-R~\citep{conneau-etal-2020-unsupervised}, mBART~\citep{mbart} and mT5~\citep{mt5}, Chinese tokens account for a very unbalanced proportion of the vocabulary.
For instance, the vocabulary of XLM-R and mBART is learned from the corpus of 100 languages, resulting in 250K subword tokens, of which Chinese tokens account for nearly 20K. 
Besides, in these models, non-Latin languages like Chinese have to learn embedding individually. 
It is difficult for them to compress the size of vocabulary by sharing subwords as Latin languages do. 
In such an overparameterized scenario, NMT models might meet a serious over-fitting issue.

To break the above two representation bottlenecks of Chinese, in this paper we introduce StrokeNet, a novel representation method for Chinese characters. 
Specifically, StrokeNet transforms each Chinese character to its corresponding stroke sequence. 
Then StrokeNet transforms each stroke to a lowercased Latin character by a predefined rule. 
With this transformation, a Chinese character is represented by a Latinzied stroke sequence, looking like an English word. 
The similar Chinese characters that share the same radical will also share the same affixes in their Latinized representation.

As StrokeNet represents Chinese as Latinized stroke sequence, we now can implement several powerful methods, which are previously not applicable to non-Latin languages, on Chinese NMT tasks. 
StrokeNet can learn subword vocabulary~\cite{sennrich2015neural} on the Latinized stroke sequence to break the learning bottleneck. 
Through this technique, StrokeNet can benefit from internal features such as radicals to enhance representation learning. 
Furthermore, StrokeNet can share subword embedding between source and target languages~\cite{press-wolf-2017-using}, leading to a significant parameter reduction and overcoming model over-fitting.
Finally, the ciphertext-based data augmentation \citep{kambhatla-etal-2022-cipherdaug}, a powerful technique in Latin languages can be added into StrokeNet to achieve performance boost in several NMT tasks.
These techniques can take full advantage of the rich internal information brought by Latinized strokes.

We conduct experiments on widely-used NIST Chinese-English, WMT17 Chinese-English and IWSLT17 Japanese-English NMT tasks. The results show that StrokeNet provides a significant performance boost over strong baselines using fewer model parameters. 
We achieve a new state-of-the-art result of 26.5 BLEU on the WMT17 Chinese-English task, with an increment of 2.1 BLEU over the scaling Transformer baseline~\cite{ott-etal-2018-scaling}.

Our main contributions are as follows:
\begin{itemize}
\item[$\bullet$] We propose StrokeNet to break the representation bottleneck of Chinese characters by capturing their rich internal features.
\end{itemize}
\begin{itemize}
\item[$\bullet$] We incorporate StrokeNet to Chinese NMT and make it possible to include the previously inapplicable methods in non-Latin languages.
\end{itemize}
\begin{itemize}
\item[$\bullet$] Our NMT models trained with StokeNet outperform strong baselines by fewer model parameters, achieving a new state-of-the-art result on the WMT17 Zh-En task.
\end{itemize}

\section{Related Work}
\begin{figure*}[t]
\centering
\includegraphics[width=0.93\textwidth]{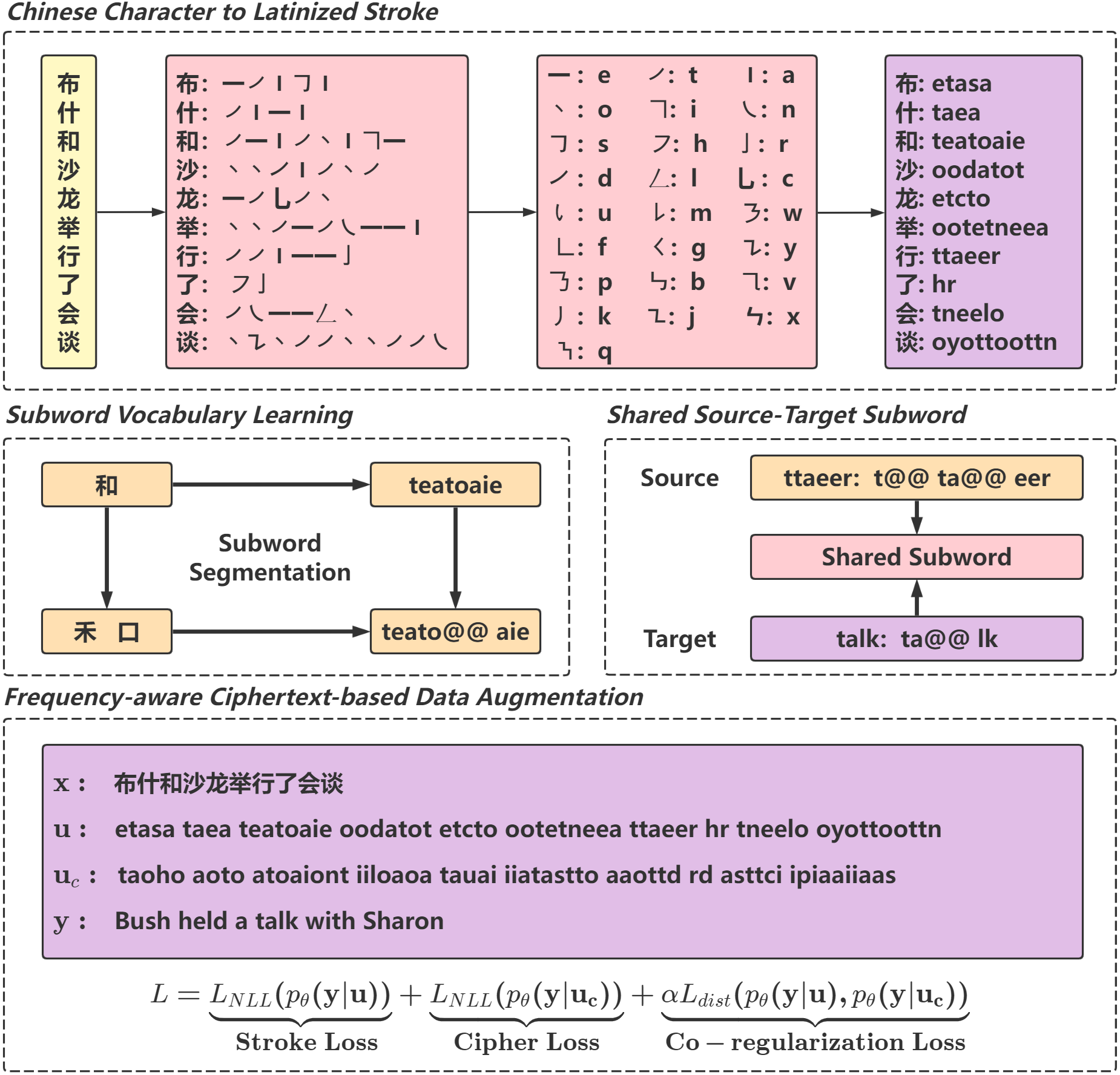}
\caption{{Overall framework of StrokeNet}. Each Chinese character is mapped to a sequence of Latin characters, like an English word. Many powerful techniques inapplicable to Chinese now can be easily applied to NMT tasks.}
\label{strokenet}
\end{figure*}

\subsection{Chinese Character Representation}
The research on Chinese character representation mainly focuses at character level and sub-character level.
Character level representation is a natural and powerful approach. 
Chinese word segmentation (CWS) is the mainstream paradigm in character level representation which cuts text into words consisting of at least one character. 
Existing research pays much attention to CWS tasks with neural network architecture.
\citet{ma2018state} use Bi-LSTMs to conduct CWS, leveraging both previous and future information in a sentence while \citet{9222029} show that self-attention network gives more competitive results.
These techniques perform well on Chinese NLP tasks but suffer from the representation learning bottleneck of the internal features.

The need to segment Chinese characters into smaller units and leverage their internal features arises in Chinese NLP tasks. 
Sub-character level representation is another promising approach. 
In Chinese, sub-characters contain internal features, such as radicals or strokes. 
Several researches focus on radical level information. 
\citet{8910369} believe that Chinese characters have a recursive structure and use treeLSTM to build hierarchical character embedding. 
\citet{DBLP:journals/corr/abs-2011-06523} leverage radical level data during training, which proves that applying radical decomposition improves Chinese-Japanese translation and performs well on translating unseen words. 

Another line focuses on stroke level information. \citet{cao2018cw2vec} propose stroke n-gram for learning Chinese character embedding with stroke n-gram information.
\citet{DBLP:journals/corr/abs-1903-00149,han2021chinese} focus on leveraging varying degrees of sub-character data, which points out that the stroke level system performs better than the ideograph level systems.
\citet{zhang-komachi-2018-neural} decompose each Chinese character into ideographs and each kind of ideograph is then decomposed into 33 kinds of predefined strokes. 

Other works try to utilize the glyph or pronunciation information in Chinese NLP tasks. 
\citet{dai-cai-2017-glyph} propose glyph-aware embedding of Chinese characters. 
\citet{sun-etal-2021-chinesebert} propose ChineseBERT, which fuses the glyph and pinyin embeddings with the original character embedding to enhance Chinese representation.
\citet{DBLP:journals/corr/abs-2106-00400} propose sub-character tokenization to encode a Chinese character into a sequence of its glyph or pronunciation and learn a new vocabulary for Chinese language model pretraining.

These researches mainly focus on learning internal features to enhance Chinese language understanding tasks.
However, the learning bottleneck still exists in Chinese NMT tasks.

\subsection{Subword Learning for NMT}
Subword learning is widely used to address the limited vocabulary problem in NMT and has been proved powerful~\citep{sennrich2015neural}. 
Several researches leverage different segmentation as augmented data or a noisy term during training. 
\citet{kudo2018subword} propose subword regularization by integrating different segmentation of words to NMT models by probability. 
\citet{provilkov2019bpe} propose the BPE-dropout technique to stochastically corrupt the segmentation procedure of BPE. 
\citet{wang2021multi} propose multi-view subword regularization to make full use of different kinds of segmentation. 
\citet{manghat2022hybrid} propose a hybrid subword segmentation algorithm to deal with out-of-vocabulary words.
\citet{DBLP:journals/corr/abs-2106-12672} propose a soft gradient-based subword tokenization algorithm to learn subword representation in data-driven fashion. 
\citet{acs2021subword} investigate how different strategies of subword pooling affect the downstream performance.

Shared embedding is a popular and powerful technique jointly used with subword learning in NMT~\citep{press-wolf-2017-using, pappas-etal-2018-beyond,liu2019latent,liu-etal-2019-shared}.
For NMT tasks in Latin languages, shared subword learning has become the de facto standard to improve the performance of NMT and compress the vocabulary size~\cite{vaswani2017attention,joshi2020spanbert,DBLP:journals/corr/abs-1901-02860}. 
This technique reduces the size of NMT models greatly without harming their performance. 
However, this technique is difficult to implement in Chinese because of the differences between Chinese and Latin languages, making it hard to break the above introduced parameter bottleneck.

\section{StrokeNet} \label{method}
To break the learning and parameter bottlenecks of Chinese character representation, we propose StrokeNet that maps a Chinese character into a Latinized sequence, and apply it to NMT tasks.
Figure~\ref{strokenet} shows the overall framework of StrokeNet.

\subsection{Chinese Character to Latinized Stroke}
\paragraph{Chinese Character to Strokes Mapping}
To learn more internal information in Chinese characters, we first need to map Chinese character to its corresponding stroke sequence.
Formally, given a Chinese sentence $\mathbf{x} = (x_1, x_2, x_3,\cdots,x_n)$, StrokeNet maps it into $\mathbf{s} = (\bm{s}_{1}, \bm{s}_{2}, \bm{s}_{3},\cdots,\bm{s}_{n})$, where $\bm{s}_{i}$ represents the corresponding stroke sequence of $x_i$. 
As shown in Figure~\ref{strokenet}, the Chinese word ``什'' can be transferred to the stroke sequence ``$\prime$ | 一 |''. 
Besides, since a small number of Chinese characters have the same sequence of strokes, we follow \citet{zhang-komachi-2018-neural} to make them distinguishable by adding a different digit at the end of the stroke sequence.
For instance, ``井'' and ``开'' have the same stroke sequence ``一 一 $\prime$ |''. In StrokeNet, the corresponding stroke sequence of ``井'' is `一 一 $\prime$ | 0'' and ``开'' is ``一 一 $\prime$ | 1''. 
Without loss of generality, we follow the implementation\footnote{\url{https://github.com/bamtercelboo/cw2vec}} of the previous work \citep{cao2018cw2vec} to define strokes, which is the most widely-used criterion consisting of 25 kinds of strokes.\footnote{\citet{DBLP:journals/corr/abs-2106-00400} represent Chinese characters into Latinized stroke sequence with 5 kinds of strokes, but do not seem to get clear improvements.}
Through stroke level representation, more internal information is easier to learn for NMT models.

\paragraph{Frequency Mapping}
To make Latin language techniques applicable in StrokeNet, we then map the stroke vocabulary, which consists of 25 kinds of strokes, to the lowercased Latin alphabet of 26 characters. 
Lexical marker is an important part of information composition. Frequent words are low-information words because they have few lexical markers~\citep{finn1977word}. 
Inspired by this information theory, we construct the mapping rule by the frequency of character occurrence. 
For instance, the Latin character ``e'' has the highest frequency of 12.7\% in English while the stroke ``一'' has the highest frequency of 27.9\% in Chinese. 
We map ``一'' to ``e'' and follow the frequency order to map the other strokes to Latin characters.
We leave the character ``z'' not mapped because we only define 25 kinds of strokes and ``z'' has the lowest frequency in English. 
Finally, we get the Latinized stroke sequence of the Chinese sentence. 
We use $\mathbf{u} = (\bm{u}_{1}, \bm{u}_{2}, \bm{u}_{3},\cdots,\bm{u}_{n})$ to represent the corresponding Latinized stroke sequence of $\mathbf{x}$.
Appendix~\ref{apd:stroke} shows the mapping dictionary.

\subsection{Application to NMT tasks}
We apply StrokeNet to Chinese NMT tasks and introduce how to combine it with other popular techniques of NMT. It is noted that the techniques are inapplicable to Chinese NMT without StrokeNet.

\paragraph{Subword Vocabulary Learning}
We use the subword vocabulary learning \citep{sennrich2015neural} technique to break the learning bottleneck of internal information.
After mapping Chinese characters to Latinized stroke sequences, characters are decomposed into smaller units. 
We conduct byte pair encoding (BPE) algorithm on the corpus of Latinized strokes. 
BPE segments Chinese characters into smaller components like subwords in English. 
During training, NMT models utilize this segmentation to learn better representation.
For instance, the character ``和'' can be cut into ``禾'' and ``口'' because its corresponding Latinized stroke sequence ``teatoaie'' can be cut into ``teato@@'' and ``aie''. 
According to \citet{li2015component}, simple characters in Chinese account for less than 20\% which cannot be split into other components. 
The others are compound characters. 
So more than 80\% of Chinese characters can benefit from our stroke-based representation.
With this advantage, we can learn stronger Chinese representation.

\paragraph{Shared Source-Target Representation}
We then use shared subword embedding \citep{press-wolf-2017-using} between Latinized strokes and English to break the parameter bottleneck. 
NMT models with shared embedding can benefit from shared source-target representation and parameter reduction. 
After transforming Chinese characters to Latinized stroke sequences, Chinese can jointly learn BPE merge operations with Latin Languages.
For example, if we cut the Latinized strokes of ``行'' into ``t@@ ta@@ eer'', and the English word ``talk'' into ``ta@@ lk'', the ``ta@@'' could be a shared subword in both Latinized Chinese and English. 
The shared source-target representation can work as a regularization term in the model training, smoothing the learning process.
Besides, the difference in parameters between NMT models with the same architecture mainly comes from the vocabulary size.
The shared subword vocabulary can also lead to a great parameter reduction.

\paragraph{Frequency-aware Ciphertext-based Data Augmentation}
As a powerful technique in NMT of Latin languages, the ciphertext-based data augmentation (CDA) is difficult to implement in Chinese NMT tasks due to the huge character list~\citep{kambhatla-etal-2022-cipherdaug}. 
StrokeNet addresses the limitation and now it can be well implemented. 
CDA is a character substitution method that replaces a character in the text with the \textit{k}th character after it in the alphabet. 
For cipher-\textit{1}, the character ``e'' is replaced by ``f'' to produce the pseudo source text and other characters follow the same rule. 
The last character ``z'' in the alphabet is replaced to ``a''. \textit{k} represents the distance between the source character and the target replaced character.

In StrokeNet, we further propose a frequency-aware ciphertext-based data augmentation (FCDA). FCDA replaces a character with the \textit{k}th character after it by the frequency order instead of alphabet order. 
For cipher-\textit{1}, the character ``e'' is replaced by ``t'' instead of ``f'' because ``e'' has the highest frequency and ``t'' has the second highest frequency.
We apply FCDA to the Latinized stroke sequence, producing the pseudo sources of the same semantic meaning and performing the multi-source learning for NMT as follows:
$$L = \underbrace{L_{NLL}(p_\theta(\mathbf{y}|\mathbf{u}))}_{\bf\small  Stroke\;Loss} + \underbrace{L_{NLL}(p_\theta(\mathbf{y}|\mathbf{u_c}))}_{ \bf \small Cipher\;Loss}$$
$$ + \underbrace{\alpha L_{dist}(p_\theta(\mathbf{y}|\mathbf{u}), p_\theta(\mathbf{y}|\mathbf{u_c}))}_{\bf\small Co-regularization \; Loss}$$
We follow \citet{kambhatla-etal-2022-cipherdaug} to minimize three losses in training, i.e., the stroke loss for the Latinized strokes, the cipher loss for the ciphered Latinized strokes, and the co-regularization loss. 
This method can reduce the impact of rare words and significantly improve performance in NMT.

\section{Evaluation}
We aim to answer the research questions through the following experiments:
\begin{itemize}
\item[$\bullet$] Can StrokeNet improve the performance of Chinese NMT tasks?
\end{itemize}
\begin{itemize}
\item[$\bullet$] Can StrokeNet reduce the scale of parameters of NMT models?
\end{itemize}

\begin{table*}[t]
\centering
\scalebox{1}{
\begin{tabular}{lcccccccc}
\toprule
\multirow{2}{*}{\bf Model} & \multicolumn{2}{c}{\bf Parameters} & \multicolumn{6}{c}{\bf Performance (BLEU)} \\
\cmidrule(lr){2-3} \cmidrule(lr){4-9}
& {\bf Total} &{\bf Emb.} &\bf Valid &\bf MT02 &\bf MT03 &\bf MT04 &\bf MT08 &\bf ALL \\ 
\midrule
Shared-Private~\scriptsize{\citep{liu-etal-2019-shared}} &63M &19M &42.6 &43.7 &42.0 &44.5 &33.8 &41.6 \\
AdvGen~\scriptsize{\citep{cheng-etal-2019-robust}} &95M &39M &47.0 &47.0 &46.5 & 47.4 &37.4 &- \\
AdvAug~\scriptsize{\citep{cheng-etal-2020-advaug}} &95M &39M &49.2 &49.0 &48.0 &48.9 & 39.6 & - \\
Manifold~\scriptsize{\citep{chen-etal-2021-manifold}} &83M &46M &49.4 &49.6 &\bf 50.3 &49.5 &39.2 &- \\
\hdashline
Vanilla &80M &36M &47.4 &47.7 &47.5 &47.7 &38.1 &45.5 \\
StrokeNet &\bf 59M &\bf 15M &\bf 49.7 &\bf 50.7 &49.8 &\bf 50.2 &\bf41.3&\bf 48.1 \\
\bottomrule
\end{tabular}}
\caption{Model parameters and performance (BLEU) on the NIST Zh-En translation task. ``Emb.'' denotes the parameters used for the embedding layer. StrokeNet provides a significant performance boost over the strong baselines with dramatically fewer model parameters.}
\label{tab:main}
\end{table*}

\subsection{Experimental Setup}
\paragraph{Data}
We conduct experiments on the NIST Zh-En and WMT17 Zh-En benchmarks. For the NIST Zh-En, the training data contains 1.25M sentence pairs. We use MT06 as the validation set and report results on MT02, MT03, MT04, and MT08 test sets, with each consists of four references. For the WMT17 Zh-En, the training data contains 20M sentence pairs. The development set is newsdev2017 and the test set is newstest2017. 
We use the scripts in Moses \citep{koehn2007moses} to tokenize and truecase the data.\footnote{\url{https://github.com/moses-smt/mosesdecoder}} 
We use \texttt{jieba}\footnote{\url{https://github.com/fxsjy/jieba}} to conduct CWS. 
Then we apply the BPE algorithm to Chinese and English separately. 
For the NIST Zh-En, we execute 30K BPE merge operations on Chinese and English separately in the baseline and 30K joint-BPE operations on Chinese and English together in StrokeNet. For the WMT17 Zh-En, we conduct 32K BPE operations on Chinese and English separately in the baseline, and 50K joint-BPE operations in StrokeNet.

Besides, to verify the validity of StrokeNet in other non-Latin languages, we also conduct experiments on the IWSLT17 Ja-En, which contains 223K training sentence pairs. The preprocessing keeps the same with the other two benchmarks.
We use \texttt{mecab}\footnote{\url{https://github.com/taku910/mecab}} to conduct Japanese word segmentation.
We make a statistic of the composition of Japanese. Chinese characters account for about 26\%. 
Japanese pseudonyms account for about 52\% and others appear to be special characters. There are 190 kinds of Japanese pseudonyms in total.
Based on this fact, we simply apply StrokeNet to those Chinese characters in Japanese after converting them to simplified Chinese.
The rest characters keep unchanged. 
The BPE merge operations are 30K for source and target separately in the baseline and 30K for joint vocabulary in StrokeNet. 
The final vocabulary sizes of these three corpora are detailed in Table \ref{tab:vocab}.

\begin{table}[t]
\centering
\scalebox{0.96}{
\begin{tabular}{ccccc}
\toprule
\multicolumn{2}{l}{\bf Model}  &\bf N. Zh-En &\bf W. Zh-En &\bf Ja-En\\
\midrule
\multirow{2}{*}{Vanilla} &Src  &40K &50K &33K \\
&Trg &30K &37K &30K \\
\hdashline
\multicolumn{2}{l}{StrokeNet} &29K &50K &28K \\
\bottomrule
\end{tabular}}
\caption{The vocabulary sizes of the three corpora. The vanilla baseline learns BPE operations separately on the source and target text. StrokeNet learns joint vocabulary and shares all embeddings of the model.}
\label{tab:vocab}
\end{table}

\paragraph{Model}
We use three kinds of Transformer architectures in the experiments. 
For the small-scale Ja-En dataset, we use the small architecture with hidden size 288, FFN size 507, 4 heads, and 5 encoder/decoder layers. 
For the medium-scale NIST Zh-En dataset, we use the base architecture with hidden size 512, FFN size 2048, 8 heads, and 6 encoder/decoder layers. 
For the large-scale WMT17 Zh-En dataset, we use the large architecture with hidden size 1024, FFN size 4096, 16 heads, and 6 encoder/decoder layers. 

\paragraph{Settings}
For training vanilla NMT models, the decoder input and output embeddings are shared. 
For the NIST Zh-En, we use Adam to train for 100K steps, with 32K max tokens per batch, the learning rate 0.0005, $\beta_1$ = 0.9, $\beta_2$ = 0.98, weight decay of 0.0001 and dropout ratio 0.3. 
We warm up the learning rate for the first 4K steps and then use the inverse square root scheduler. 
For the WMT17 Zh-En, the number of max tokens per batch is 288K and others keep the same with the NIST Zh-En. 
For the Ja-En, we train for 100K steps with 16K max tokens per batch and the learning rate is 0.0003 with a cosine scheduler. 

For training NMT models with StrokeNet, each Chinese character is mapped to the corresponding Latinized stroke sequence.
Joint vocabulary is learned from both source and target texts together. 
During training, all the embeddings and softmax weights are shared. 
For the Ja-En, NIST Zh-En, and WMT17 Zh-En, the max tokens per batch are 16K, 96K, and 192K respectively. 
The rest hyperparameters keep the same with the vanilla models.

During testing, for the NIST Zh-En and Ja-En, we generate with length penalty 1.0 and beam size 5.
For the WMT17 Zh-En, we generate with length penalty 1.4 and beam size 5.
The BLEU scores are evaluated with {\it multi-bleu.perl} provided by Moses~\citep{koehn2007moses}.
We use the checkpoint with the best validation BLEU for testing.

\subsection{Main Results}
\paragraph{Parameter Reduction}
To verify whether StrokeNet can reduce the parameter of NMT models, we look into the parameter and vocabulary size in each experiment. 
Table~\ref{tab:vocab} and Table~\ref{tab:main} show the parameters and vocabulary sizes in all experiments. 
StrokeNet decreases the vocabulary size and the parameters obviously. 
For the NIST Zh-En, the embedding layer parameters are 15M, which is the smallest. 
The model parameters of StrokeNet are 59M, while the vanilla baseline and previous methods are over 80M generally.
Compared with the prior work \citep{chen-etal-2021-manifold} with competitive performance, StrokeNet has a 24M parameter reduction.
This parameter reduction can greatly decrease the pressure of computational complexity while achieving more competitive results.

For the WMT17 Zh-En and Ja-En, we list their vocabulary sizes in Table~\ref{tab:vocab}. 
StrokeNet gains 37K, and 35K vocabulary reduction respectively on these two corpora.
With the smaller vocabulary, StrokeNet reduces the representation redundancy in Chinese characters and learns shared representation.
As the parameter reduction of NMT models mainly comes from the vocabulary size reduction, StrokeNet on these two datasets also has an obvious parameter reduction. 
These results show that StrokeNet breaks the parameter bottleneck, which can alleviate the over-fitting problem.

\paragraph{Performance Boost}

\begin{table}[t]
\centering
\scalebox{0.95}{
\begin{tabular}{l r r c}
\toprule
{\bf Model} &\bf Ja-En &\bf W. Zh-En \\ \midrule
Shared-Private~\scriptsize{\citep{liu-etal-2019-shared}} &12.4 &- \\
Norm~\scriptsize{\citep{liu-etal-2020-norm}} &- & 25.3 \\
Prior~\scriptsize{\citep{chen2021integrating}} &- & 25.5 \\
\hdashline
{Vanilla} &12.0 &24.4  \\
{StrokeNet} &\bf 13.1 &\bf26.5   \\
\bottomrule
\end{tabular}}
\caption{Performance (BLEU) on the Ja-En and large-scale WMT17 Zh-En benchmarks.}
\label{tab:a}
\end{table}

To verify whether StrokeNet can improve performance on Chinese NMT tasks, we look into the results of the three benchmarks.
Table \ref {tab:main} shows the translation performance of the validation and test sets on the NIST Zh-En benchmark. StrokeNet obtains a BLEU of 48.1 on the collection of all test sets, an improvement of 2.6 BLEU over the vanilla baseline.
We also see improvements over prior work \citep{chen-etal-2021-manifold} on every subset except MT03. 

Table \ref{tab:a} shows the translation performance on the test sets for the Ja-En and large-scale WMT17 Zh-En. 
For the Ja-En, StrokeNet improves translation quality by 1.1 BLEU over the vanilla baseline and 0.7 BLEU over the prior work \citep{liu-etal-2019-shared}. 
Although there are only about 26\% Chinese characters in Japanese, StrokeNet can still gain 1.1 BLEU improvement on the Ja-En task.
For the WMT17 Zh-En, StrokeNet achieves a new state-of-the-art result of 26.5 BLEU, obtaining +2.1 BLEU over the vanilla baseline, +1.2 BLEU over \citet{liu-etal-2020-norm}, and +1.0 BLEU over \citet{chen2021integrating}.
In the future, we will further enhance the NMT models with pretrained knowledge~\citep{liu-etal-2021-complementarity-pre,liu-etal-2021-copying}.
The above results confirm the effectiveness of StrokeNet, which can be applied to different scales of training data and different languages.

\section{Analysis}

\paragraph{Effect of BPE Merge Operations} \label{merge}
\begin{figure}[t]
\centering
\includegraphics[width=1.0\columnwidth]{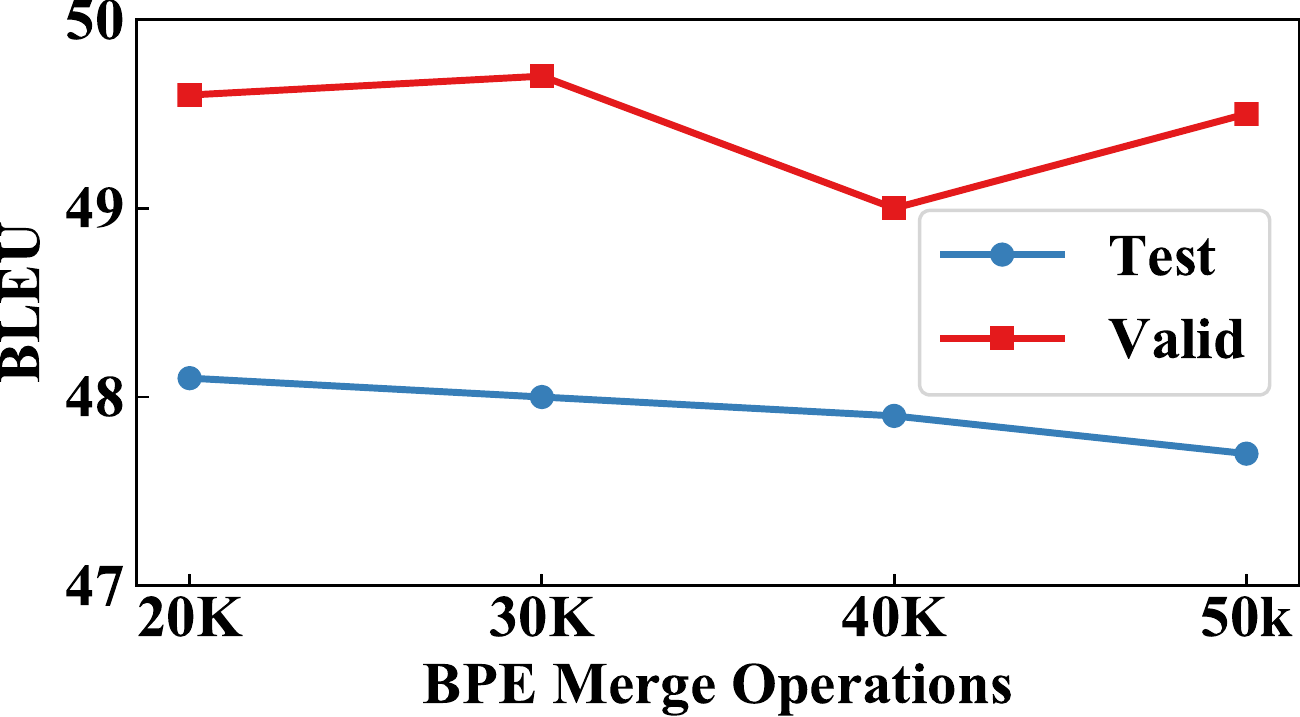}
\caption{BLEU on the valid and test sets for different numbers of BPE merge operations.}
\label{bpe}
\end{figure}

StrokeNet benefits from the subword modeling technique. 
To explore how it works in StrokeNet, we conduct experiments on the NIST Zh-En benchmark, applying different numbers of BPE merge operations. 
We conduct experiments on 20K, 30K, 40K, and 50K merge operations. 
Results are detailed in Figure \ref{bpe}. 
30K merge operations appear to be the best choice. 
For the validation set, the translation performance reaches the highest at 30K. For the test set, it gradually decreases as the number of merge operations increases. 
We see variations of less than 0.7 BLEU in the dev set and less than 0.4 BLEU in the test set as the number of BPE merge operations changes. 
And large improvements over the vanilla baseline are observed regardless of the number of BPE merge operations with at least +2.2 BLEU for both the validation and test sets.
The results show that StrokeNet is robust to the number of BPE merge operations.

\paragraph{Learning Curves}

\begin{figure}[t] 
\centering
\includegraphics[width=1.0\columnwidth]{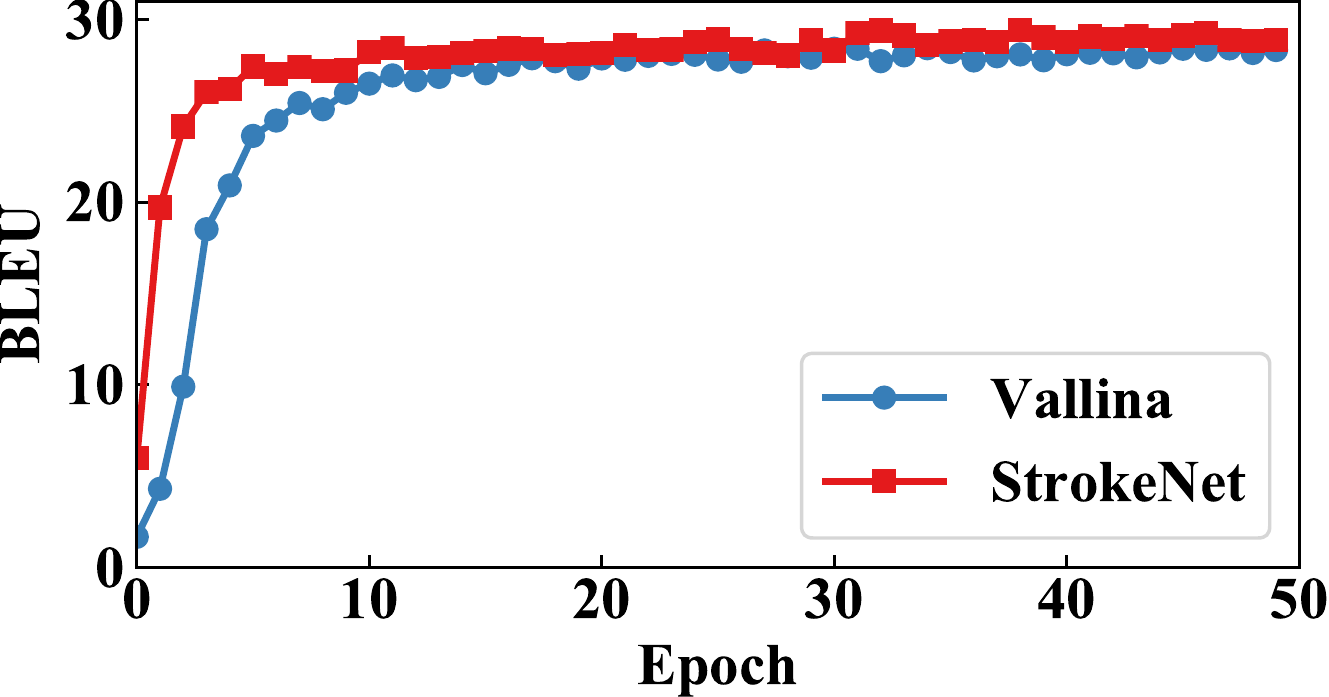}
\caption{Learning curves for the vanilla baseline and StrokeNet.} 
\label{learning curves} 
\end{figure}
To explore how StrokeNet performs better than the vanilla baseline, we draw the learning curves during training. 
We train StrokeNet and the vanilla baseline for both 50 epochs on the NIST Zh-En. 
The validation BLEU during training is shown in Figure \ref{learning curves}.\footnote{The BLEU here is calculated with one reference.} 
In StrokeNet, the BLEU on the validation set rises faster in the early period and finally achieves higher BLEU than the vanilla baseline. 
With the Latinized stroke level representation and the application of powerful techniques in Latin languages, NMT models with StrokeNet learn faster and better than vanilla models.
The rich internal information in Chinese characters becomes readily available for StrokeNet to learn.
The results prove that StrokeNet successfully breaks the representation learning bottleneck, showing its positive effects on Chinese NMT tasks.

\paragraph{Effect of Data Scale}
\begin{figure}[tp]
\centering
\includegraphics[width=1.0\columnwidth]{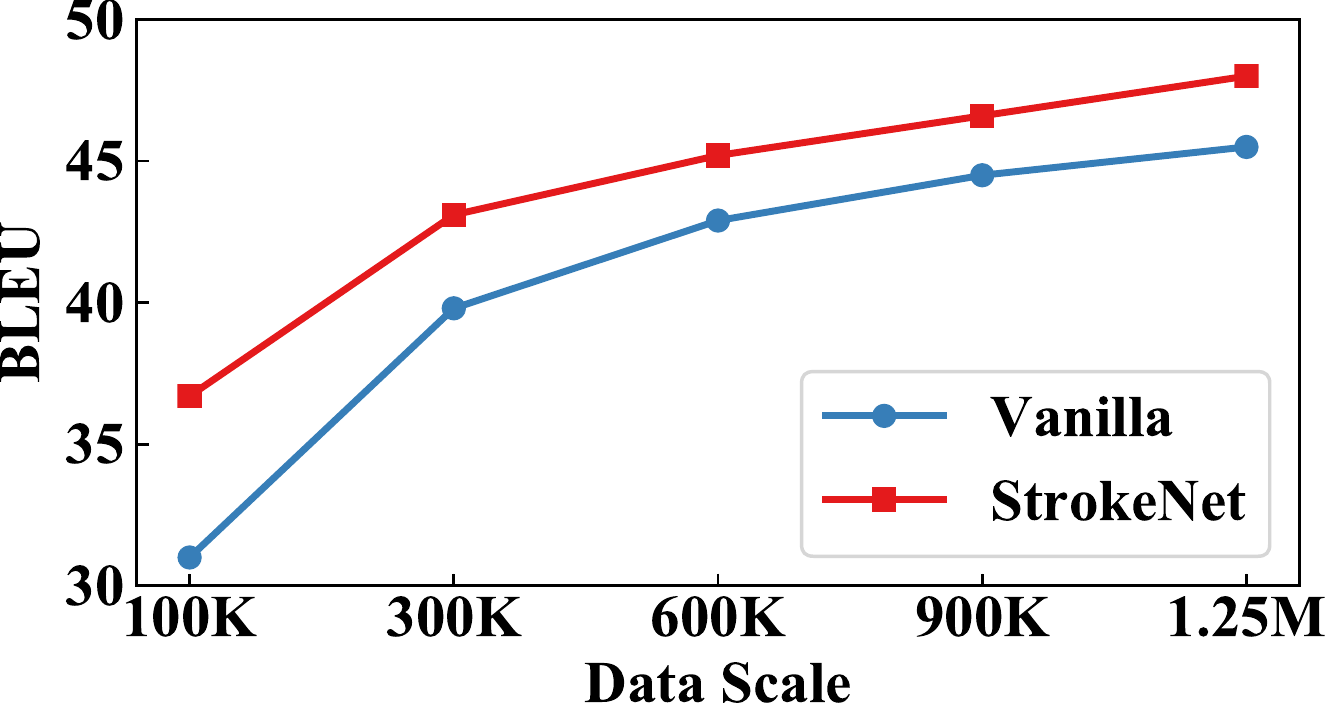}
\caption{BLEU on the NIST Zh-En for different scales of training data. We randomly extract the data subsets from the 1.25M sentence pairs.}
\label{data scale}
\end{figure}

To further illustrate the effects of different data scales in StrokeNet, we randomly extract four subsets from the original 1.25M source sentence pairs in the NIST Zh-En. 
We control their size to be 100k, 300k, 600k, and 900k. 
The results on the test set are given in Figure \ref{data scale}. 
With each data scale, StrokeNet yields large improvements over the vanilla baseline by 2.1-5.7 BLEU. 
Furthermore, with the decrement of data scale, the performance margin between StrokeNet and the vanilla baseline becomes larger. 
In particular, the improvement is much larger under the 100K setting (+5.7 BLEU) than that under the 1.25M setting (+2.6 BLEU). 
Besides, we also see great improvements on the low-scale Ja-En and large-scale WMT17 Zh-En. 
StrokeNet is proved powerful on NMT tasks of varying data sizes.

\paragraph{Ablation Analysis}
To explore which part of StrokeNet makes a difference, we conduct several ablation experiments on the NIST Zh-En benchmark.
First, we explore the effect of the frequency mapping technique, which is inspired by the information theory~\citep{finn1977word}. 
The theory states that more frequent words are lower information words because they have fewer lexical markers. 
We compare StrokeNet with frequency mapping to StrokeNet with randomly mapping, which means that each kind of stroke is mapped to a unique Latin character randomly. 
Table \ref{tab:ablation} shows that the performance of StrokeNet with mapping randomly is 0.6 BLEU worse than frequency mapping, which proves that our application of the information theory is reasonable. Mapping units in two languages by their frequency reduces information loss.

To further explore how frequency mapping improves performance, we conduct statistics of shared subwords in the NIST Zh-En data obtained using frequency mapping and random mapping with the same BPE merge operations. 
Table \ref{tab:shared-subword} gives the results. 
Ratio refers to the ratio of shared subwords over the whole subword in the training data. 
Length refers to the weighted average of the length of shared subwords. 
BPE is also a mapping algorithm based on frequency and can benefit from the proposed frequency mapping. 
The results show that frequency mapping produces more shared subwords and longer subword units between the Latinized stroke sequences and English, resulting in shorter sequences, which can lead to stronger memorization in Transformer models \citep{DBLP:journals/corr/abs-2110-02782}, and thus better translation quality.

\begin{table}[t]
\centering
\scalebox{0.96}{
\begin{tabular}{l c c}
\toprule
{\bf Model} & \bf Emb.  &\bf BLEU \\ \midrule
Vanilla &36M &45.5 \\
\hdashline
{StrokeNet} w/ Rand. Mapping &15M &47.5 \\
{StrokeNet} w/ Freq. Mapping &15M &48.1 \\
\hdashline
{StrokeNet} w/o Freq. CDA &15M &45.9 \\
{StrokeNet} w/o Shared Voc. &21M &47.7 \\
\bottomrule
\end{tabular}}
\caption{Performance (BLEU) of different model variants on the NIST Zh-En benchmark.}
\label{tab:ablation}
\end{table}

\begin{table}[t]
\centering
\scalebox{0.96}{
\begin{tabular}{l c c}
\toprule
{\bf Model} & \bf Ratio  &\bf Length \\ \midrule
{StrokeNet} w/ Rand. Mapping &37.9 &5.7 \\
{StrokeNet} w/ Freq. Mapping &39.1 &5.8 \\
\bottomrule
\end{tabular}}
\caption{Statistics of shared subwords in the NIST Zh-En data obtained using frequency mapping and random mapping with the same 30K BPE merge operations.}
\label{tab:shared-subword}
\end{table}

Second, we explore the effect of FCDA. 
We conduct experiments on StrokeNet without FCDA and keep the other settings unchanged.
The result in Table \ref{tab:ablation} still yields an improvement over the vanilla baseline.
StrokeNet enables this strong data augmentation technique in Latin languages to be implemented for Chinese NMT tasks. 

Finally, we explore the effect of shared vocabulary and embedding. 
We use this technique to achieve shared representation and parameter reduction. 
We conduct experiments in StrokeNet without sharing vocabulary. The results are detailed in Table \ref{tab:ablation}.
Without sharing vocabulary, the performance decreases by 0.4 BLEU but still gains large improvements over the vanilla baseline.
The parameters of StrokeNet are 6M fewer than StrokeNet without sharing vocabulary, which is consistent with intuition.
Through sharing vocabulary, StrokeNet achieves parameter reduction and better performance by learning shared source-target representation. 
This means that the shared subword learning technique works well in StrokeNet.

\paragraph{Translation of Word of Different Frequency}

\begin{table}[t]
\centering
\scalebox{1}{
\begin{tabular}{l c c c c}
\toprule
{\bf Model} &\bf Low &\bf Medium &\bf High &\bf Total\\ 
\midrule
Vanilla &38.9  &47.0  &63.2 &59.5 \\
StrokeNet &39.2  &48.5  &64.2 &60.6 \\
\bottomrule
\end{tabular}}
\caption{Prediction accuracy for words of different frequencies. StrokeNet performs well on medium- and high-frequency words.}
\label{tab:freq}
\end{table}

To explore the translation quality difference between the vanilla baseline and StrokeNet, we conduct an analysis by comparing accuracy on different frequency words in the test set on the NIST Zh-En benchmark.
The frequency of words is based on the training set. 
As shown in Table \ref{tab:freq}, StrokeNet achieves pretty good translation accuracy on medium and high-frequency words. 
For words of medium frequency between 200 and 2,000, StrokeNet achieves 48.5 and shows an improvement of 1.5 BLEU over the vanilla baseline. 
For words of high frequency over 2,000, it achieves 64.2 while the baseline achieves only 63.2. 
Words of low frequency, also known as rare words, still get an improvement of 0.3 over the vanilla baseline. 
For all the words, StrokeNet improves the prediction accuracy from 59.5 to 60.6. 
The results show that representation learning has been improved by learning more internal features through stroke modeling.

\section{Conclusion}
In this paper we introduce StrokeNet, a novel technique for Chinese NMT tasks using Latinized stroke sequence of Chinese characters. 
StrokeNet breaks the representation learning bottleneck and the parameter bottleneck in Chinese NMT tasks, which requires no external data and significantly outperforms several strong prior works. 
We show that representing Chinese characters in stroke level works well on NMT tasks to bring more internal structure information. 
We demonstrate that it is possible to implement popular and powerful techniques designed for Latin languages in Chinese NMT tasks. 
We conduct several analyses on the effects of these Latin language techniques, proving they bring an obvious performance boost to StrokeNet. 
Overall, StrokeNet is a simple and effective approach for Chinese NMT tasks and yields strong results in both high-source and low-source settings.
Future work includes applying StrokeNet to other language generation tasks~\citep{liu2021understanding}.

\section*{Limitations}
\begin{CJK}{UTF8}{bsmi}
Challenges remain in StrokeNet. As shown in Table \ref{tab:freq}, even with the best BPE merge operations, the translation accuracy of low-frequency words gains a minor boost over the baseline by just 0.3, which is not as good as middle and high-frequency words.
We speculate that low-frequency Chinese characters might be hurt when they are cut into subwords. 
For example, the low-frequency Chinese character ``劑 (medicament)'', whose corresponding Latinized stroke sequence is ``oeotasttmntaeear'', is segmented into ``oeot@@ a@@ stt@@ m@@ n@@ ta@@ eea@@ r''. It is too chopped up and its semantic information becomes incomplete. 
How to handle this kind of segmentation and improve the translation quality of low-frequency Chinese characters remains to be explored.
\end{CJK}

\section*{Acknowledgments}
This work was supported in part by the National Natural Science Foundation of China (Grant No. 62206076) and Shenzhen College Stability Support Plan (Grant No. GXWD20220811173340003 and GXWD20220817123150002).
The computational resources were supported by Education Center of Experiments and Innovations at Harbin Institute of Technology, Shenzhen.
We would like to thank the anonymous reviewers and meta-reviewer for their insightful comments and suggestions.

\bibliography{emnlp2022}
\bibliographystyle{acl_natbib}
\appendix

\newpage
\section{Appendix}
\label{sec:appendix}
\subsection{Statistical Data of Character and Stroke Frequencies}
\label{apd:stroke}
\begin{figure}[ht]
\centering
\includegraphics[scale=0.181]{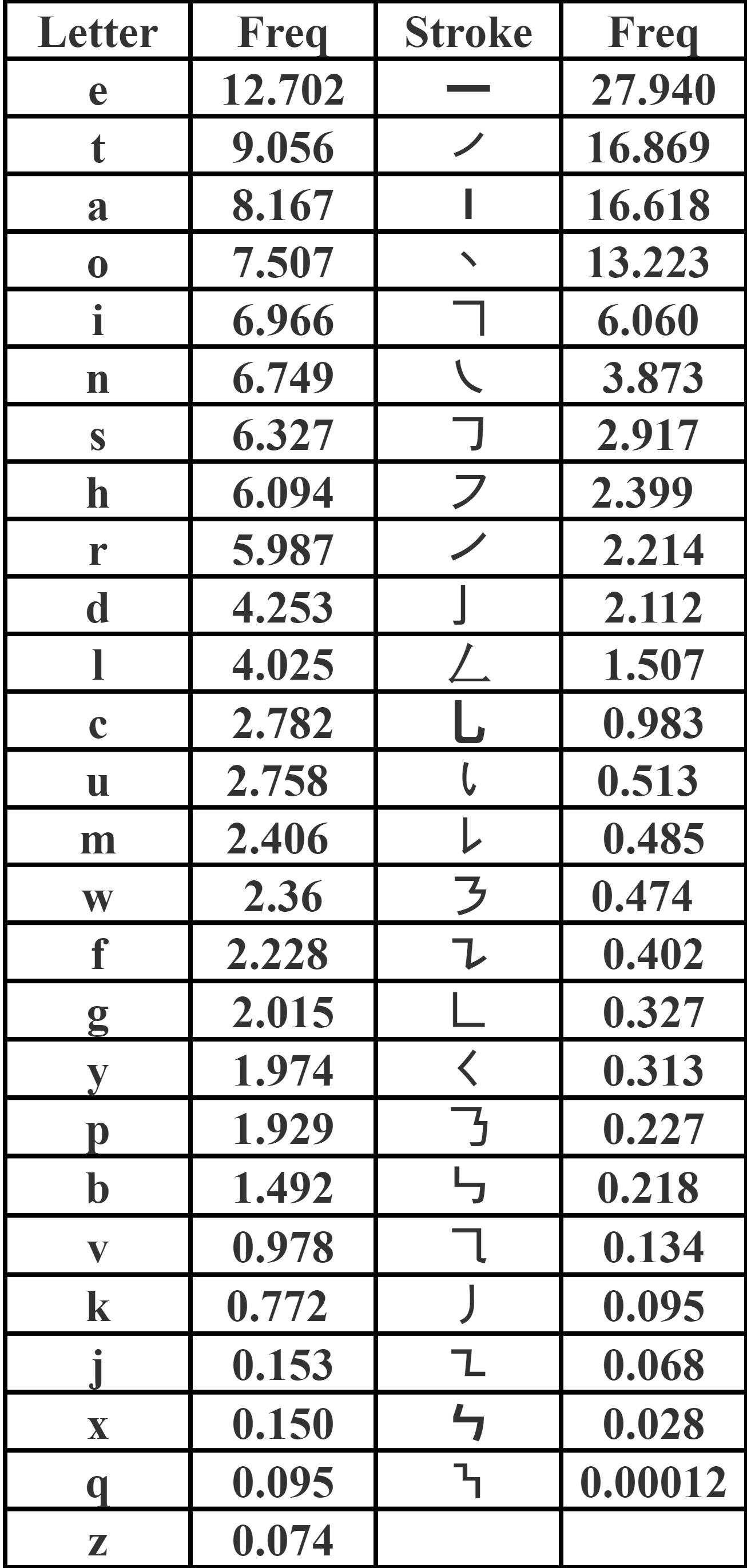}
\caption{Frequencies of Latin lowercased characters and Chinese strokes.}
\label{freq}
\end{figure}

Figure \ref{freq} shows the frequency of occurrence of each lowercased Latin character and each Chinese stroke. 
The frequency of each lowercased Latin character is from Wikipedia.\footnote{\url{https://en.wikipedia.org/wiki/Letter_frequency}}
The frequency of each stroke is from WMT17 Zh-En data, which contains 20M Chinese sentences. 
We order them by frequency and each stroke is mapped to the Latin lowercased character in the same row. ``z'' has no corresponding stroke because we only define 25 kinds of strokes and it has the minimum frequency.

\end{CJK*}
\end{document}